\documentclass[10pt, a4paper]{article}
\usepackage[table,xcdraw]{xcolor}
\usepackage[]{lrec-coling2024} % this is the new style

\usepackage{graphicx}
\usepackage{array}
\usepackage{bm}
\usepackage{multirow}
\usepackage{float}%%%%提供浮动体的[H]选项，进而取消浮动
\usepackage{subfigure} %插入多图时用子图显示的宏包
\usepackage{booktabs}
\usepackage{amsmath}
\usepackage{amsfonts}
\usepackage{makecell}
\usepackage{enumitem}
\usepackage{color}
\usepackage[normalem]{ulem}

\title{A Logical Pattern Memory Pre-trained Model for Entailment Tree Generation}

\name{Li Yuan$^{1}$, Yi Cai$^{1,2}$, Haopeng Ren$^{1}$, Jiexin Wang$^{1,*}$\thanks{$^{*}$Corresponding authors}}

\address{$^{1}$1School of Software Engineering, South China University of Technology \\
         $^{2}$Key Laboratory of Big Data and Intelligent Robot \\(South China University
of Technology) Ministry of Education\\
        jiexinwang@scut.edu.cn
}

\abstract{
Generating coherent and credible explanations remains a significant challenge in the field of AI. In recent years, researchers have delved into the utilization of entailment trees to depict explanations, which exhibits a reasoning process of how a hypothesis is deduced from the supporting facts. However, existing models often overlook the importance of generating intermediate conclusions with logical consistency from the given facts, leading to inaccurate conclusions and undermining the overall credibility of entailment trees. To address this limitation, we propose the logical pattern memory pre-trained model (LMPM). LMPM incorporates an external memory structure to learn and store the latent representations of logical patterns, which aids in generating logically consistent conclusions. 
Furthermore, to mitigate the influence of logically irrelevant domain knowledge in the Wikipedia-based data, we introduce an entity abstraction approach to construct the dataset for pre-training LMPM.
The experimental results highlight the effectiveness of our approach in improving the quality of entailment tree generation. By leveraging logical entailment patterns, our model produces more coherent and reasonable conclusions that closely align with the underlying premises. Code and Data are released at \url{https://github.com/YuanLi95/T5-LMPM}
 \\ \newline \Keywords{Entailment Tree, Memory Network, Logical Pattern}}

\begin{document}

\maketitleabstract

\section{Introduction}

\begin{figure}    
  \centering
    \includegraphics[scale=0.63]{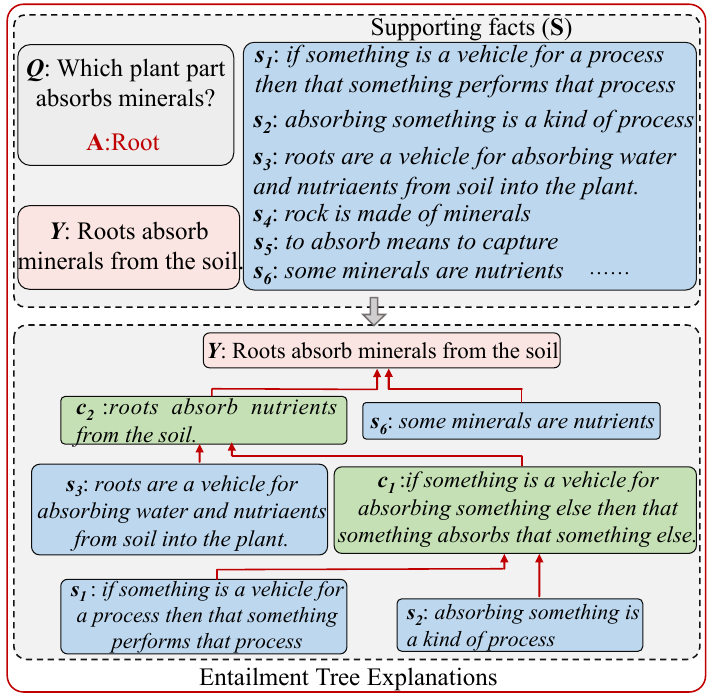}
  %\caption{In the entailment tree generation task, given a hypothesis and multiple supporting facts, the model generates an entailment tree (Figure~\ref{FIG:1} (a)), including both the tree structure and intermediate conclusions (\textbf{$c_1$} and \textbf{$c_2$}), and current methods (Figure~\ref{FIG:1} (b), (c), and (d)) for entailment tree generation tasks.}
  % \caption{An entailment tree generation example (Figure~\ref{FIG:1} (a)) and current methods for solving the task of entailment tree generation(Figure~\ref{FIG:1} (b), (c), and (d)).}
  \vspace{-0.4cm}
    \caption{Illustration of Entailment Tree Generation. The top half presents the inputs, while the generated entailment tree is depicted in the bottom half. The tree consists of a hypothesis (pink), premises (blue), and generated intermediate conclusions (green).}
      
  \label{FIG:1}
  \vspace{-0.4cm}
\end{figure}

Generating coherent and credible explanations poses a significant challenge in AI, and addressing it represents a giant leap towards the goal of building reliable reasoning systems \cite{10.1016/j.inffus.2019.12.012}. 
The reasoning chain has traditionally served as the primary representation for constructing reasonable explanations \cite{DeYoung2020,Tafjord2021}. However, in recent years, researchers have explored the utilization of entailment trees for explanation generation \cite{Dalvi2021,Yang2022,Hong2022}. 
% Recently, generating explanations in the form of entailment tree based on natural language has interested researchers \cite{Dalvi2021}.
% , which requires compositional generalization to new samples \cite{ouyang2022training}.   
As shown in Figure ~\ref{FIG:1}, the task of entailment tree generation can be defined as follows: given a hypothesis (summarizing from a question+answer pair) and a set of supporting facts, the goal is to derive an entailment tree where each non-leaf node is an intermediate conclusion generated from its children. By providing valid entailment trees, users can develop a better understanding of the reasoning process and acquire reliable information for effective decision.
% The early work \cite{Dalvi2021} propose to generate the full entailment trees in one step by an end-to-end generative model, as shown in Figure ~\ref{FIG:1}(b). However, such single-step method suffers from the unreliable explanation of the hypothesis, due to not explicitly constraining the validity of each step and the tree structure. 

The early work by \citet{Dalvi2021} proposes a single-step method for generating entailment trees. However, subsequent research conducted by \citet{Hong2022} has highlighted that such an approach often yields unreliable explanations for the hypothesis. To tackle this limitation, several recent studies employing multi-step generation methods have been introduced \cite{Yang2022, NevesRibeiro2022, Hong2022,liu-etal-2022-rlet}. These methods have demonstrated impressive performance by iteratively selecting premises and generating intermediate conclusions to construct the entire tree.
% as shown in Figure~\ref{FIG:1}(c) or (d). %\footnote{Both verifier-based and controller-based methods in the figure are belong to the iterative approach.}. 
% However, most existing iterative methods pay little attention to the logical patterns between premises and conclusions, despite the fact that they play an important role in generating logically consistent conclusions.% , as demonstrated by \citet{Bostrom2021}.
% \footnote{Note that even though \citet{Hong2022} utilizes the logical patterns to some extent, their method brings two more issues, which we discuss in the next section.} 
Nonetheless, a significant drawback of many existing iterative methods lies in their limited attention to the logical patterns between premises and conclusions. As a result, they might generate intermediate conclusions that lack logical consistency and conflict with the premises.

% For instance, the example with the correct conclusion being \emph{"precipitation and  infiltration are different stag in the water cycle process"} in Figure~\ref{FIG:2} (a)\footnote{The example is located in blue box of the top sub-figure.}, these methods might generate wrong conclusions like \emph{"infiltration is precipitation of water cycle process"}. 

% We consider the deduction process as a logical pattern, as the gray box.  If a  model learns similar correlation patterns and uses the pattern to assist the generation process will improve the generated conclusion logically consistent.
% Hong et al. (2022) indicate learning correlation
% logical patterns will improve the generated conclusion logically consistent, followed by training the T5 via synthetic Wikipedia data (green boxes).

\begin{figure}    
  \centering
    \includegraphics[scale=0.5]{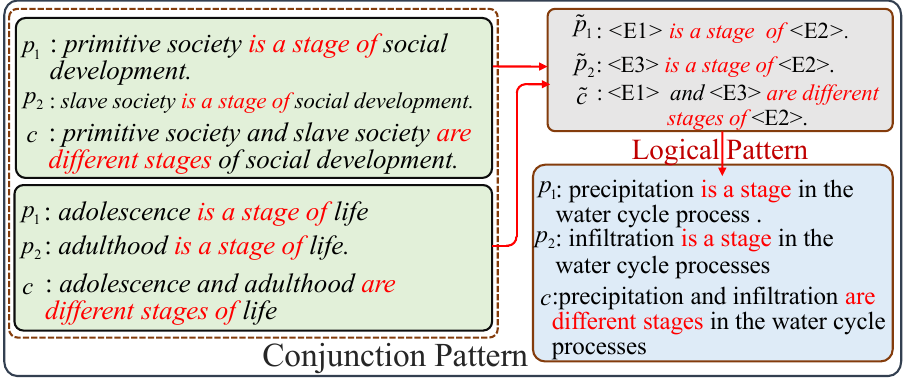}
    % \vspace{-1.9em}
     \vspace{-0.8cm}
  \caption{Examples of Logical Patterns. Each box represents a single-step deductive process. The texts within the green and blue boxes are extracted from the Wikipedia-based synthetic dataset and the entailment tree corpus, respectively. The text within the gray box is the logical patterns obtained through entity abstraction, where <E> denotes the special token <extra\_id\_ > reserved in T5.}
    
  \label{FIG:2}
  \vspace{-0.6cm}:
\end{figure}

Among the iterative methods, METGEN \cite{Hong2022} stands out for its attempt to integrate logical patterns into the tree generation process. However, it primarily focuses on the module responsible for premise selection, with relatively limited attention to effectively capturing and learning logical patterns. METGEN's strategy for integrating logical patterns involves training the language model on a synthetic dataset constructed from Wikipedia. This dataset exhibits logical regularities, as illustrated by the green boxes in Figure~\ref{FIG:2}. 
We argue that this approach is insufficient for comprehensively learning the intricacies of logical features. On one side, during training, the language model is forced to learn both logical patterns and textual features, which dilutes its focus on capturing logical patterns. Furthermore, using Wikipedia data as a training resource introduces a  considerable amount of domain-specific information that isn't logically relevant. For instance, in Figure~\ref{FIG:2}, when METGEN leverages data containing logical regularities (green boxes) to learn corresponding logical patterns (gray box) that capture generalizable logical relationships between entities, it inadvertently incorporates specific entity terms and domain-specific knowledge (e.g., \emph{primitive society} and \emph{social development}). This inclusion of specific information in the logical pattern training data poses challenges for METGEN in effectively acquiring logical patterns, which hinders its ability to generalize the reasoning process and identify similar relationships across various instances.

\begin{figure*}    
  \centering
    \includegraphics[scale=0.65]{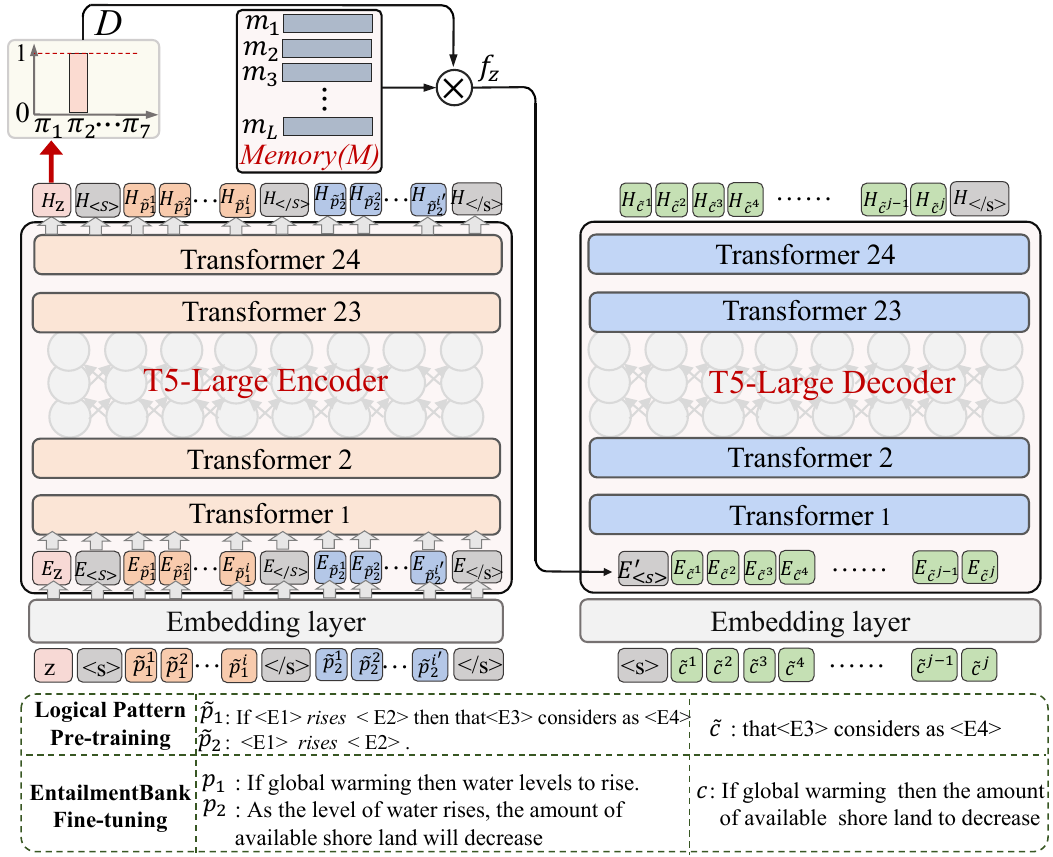}
    \vspace{-0.3cm}
  \caption{The overall architecture of LMPM} %for the generation of intermediate conclusions using the select premises.}
  \label{FIG:3}
\vspace{-0.3cm}
\end{figure*}

In this paper, we propose the \textbf{L}ogical Pattern \textbf{M}emory \textbf{P}re-trained \textbf{M}odel (LMPM) to address the aforementioned limitations, as depicted in Figure \ref{FIG:3}. Our primary objective is to effectively exploit logical patterns to generate logically consistent conclusions. To address the limitation regarding the inadequate focus on capturing logical patterns, we introduce an external memory component that operates independently of the language model. This memory module is designed to learn and store latent logical patterns between premises and conclusions, enabling the model to better capture and utilize logical relationships.
Furthermore, to mitigate the influence of logically irrelevant domain knowledge in the Wikipedia-based data, we adopt an entity abstraction approach, as illustrated in Figure \ref{FIG:2}. This involves abstracting entities from the Wikipedia text, resulting in an entity-abstract dataset utilized for pre-training our LMPM. %By incorporating this entity abstraction approach, the model becomes less affected by irrelevant knowledge and can be effectively trained with a reduced amount of data. 
By leveraging the abstraction data, the model can uncover fundamental logical relationships, thereby enhancing its reasoning capabilities while reducing the impact of irrelevant knowledge and necessitating less training data. In the experiments, we integrate LMPM with METGEN, 
% we integrate LMPM with existing iterative methods,
simply using it to generate the intermediate conclusions of logical consistency.
To sum up, we make the following contributions in this work:
\begin{itemize}[itemsep=2pt,topsep=0pt,parsep=0pt,leftmargin=10pt]
\item We introduce a logical pattern memory pre-trained model (LMPM), which facilitates the generation of logically consistent intermediate conclusions during entailment steps. LMPM employs an external memory to learn and retain latent logical patterns between premises and conclusions. This mechanism significantly enhances the language model's capacity to capture and utilize logical patterns effectively.

\item We propose to pre-train the LMPM model via a constructed entity-abstract dataset, explicitly guiding the model to learn representations of latent logical patterns. This approach mitigates the influence of irrelevant domain knowledge in the original Wikipedia data and enables the model to be well-trained with less data.

% We propose to pre-train the model via a constructed entity-abstract dataset, which explicitly forces the model to learn representations of latent logical patterns. This approach reduces the influence of irrelevant domain knowledge present in the original Wikipedia data and allows the model to be well-trained with fewer data.

%We use a special token to abstract the entities of synthesized Wiki data, which enhances the ability of memory structure to capture latent logic patterns. Profiting from the entity abstracting to reduce the logically irrelevant features, our proposed method can be reduce the training data having similar logical patterns.

\item We assess our proposed method using both automatic and human evaluations on entailment tree generation datasets. The results demonstrate that LMPM outperforms strong baseline models in most cases and provides effective representations of logical patterns.
\end{itemize}

\section{Related Work} 
\subsection{Entailment Tree Generation}
Question answering (QA) has garnered significant research interest, with recent works dedicated to explaining QA as a new departure for responsible AI systems \cite{ye2020teaching,PENG2020426,Zhang2021,9699028, mavi2022survey,10175600}. Nonetheless, most of these efforts have primarily focused on identifying relevant facts for questions and answers \cite{jansen2019textgraphs,yadav-etal-2019-quick,YUAN202210304}, rather than constructing valid reasoning paths. While some studies have explored the generation of explanation sentences leading to answers \cite{camburu2018snli, rajani-etal-2019-explain,wu2023generating}, these explanations are often unreliable as they are not guided by explicit facts \cite{Bostrom2021}. To overcome these limitations, \citet{Dalvi2021} introduce EntailmentBank, a dataset specifically designed for the task of entailment tree generation. Each multi-step entailment tree in EntailmentBank serves as an explanation, clearly demonstrating the reasoning process behind a hypothesis based on a set of supporting facts, as shown in Figure ~\ref{FIG:1}. 

In addition to introducing the EntailmentBank, \citet{Dalvi2021} propose the EntailmentWriter model to tackle the task. This model utilizes an end-to-end language model to generate complete entailment trees in a single step. However, a drawback of this approach is that it does not explicitly constrain the validity of every single step, leading to potentially unreliable explanations for the given hypothesis. 
To address the limitations of single-step methods, several recent works \cite{NevesRibeiro2022, Hong2022,liu-etal-2022-rlet} have presented multi-step generation approaches, which iteratively select premise facts and generate intermediate conclusions. For instance, \citet{NevesRibeiro2022} have devised an iterative retrieval-generation framework (IRGR) that leads to enhanced retrieval results in the context of Task 3. \citet{liu-etal-2022-rlet} have introduced a Reinforcement Learning-based Entailment Tree generation framework (RLET), which is trained by utilizing cumulative signals across the entire tree. However, such methods suffer from a coupling issue as the premise selection and conclusion generation are performed simultaneously by the T5 model. On the other hand, \citet{Hong2022} introduced the METGEN model, which is a controller-based model consisting of proprietary modules for selection and generation.
METGEN primarily focuses on the controller module, which is responsible for selecting the premises, followed by an independent T5 to generate the conclusion. Unlike other works that completely overlook logical relations, METGEN slightly improves its ability of logical summarizing by training the model with a Wikipedia-based synthetic dataset. 

In contrast to the aforementioned approaches, we focus on improving the model's ability to generate logically consistent conclusions in the entailment steps by leveraging logical patterns. We underscore the significance of integrating logical patterns into the reasoning process. 
To achieve this, we employ an external memory to both learn and store the features of logical patterns, facilitating the generation of logically consistent conclusions. Moreover, our model can be adapted to enhance existing methods like METGE, leveraging the advancements of LMPM in effectively incorporating logical patterns into the entailment tree generation.

% Unlike the above approaches, we focus on improving the model's ability to generate logically consistent conclusions in the entailment steps by using logical patterns to assist language models.
% % offering logical pattern features. 
% % better logical pattern features. 
% We emphasize the importance of the utilization of logical patterns in reasoning and use an external memory to learn and store the features of logical patterns, which can be easily reused by other approaches (e.g., NLProofs, METGEN) for the entailment tree generation.
\vspace{-0.5cm}
\subsection{Memory Networks}
\citet{weston2015memory} first propose the memory network, which employs memory components to store scene information, thereby enabling the network to maintain long-term memory. Since then, memory networks have found widespread application in various natural language processing tasks. They have demonstrated effectiveness in storing multi-hop or long-term information across contexts, notably in applications such as dialogue systems \cite{wang-etal-2020-dual,chen2023learning}, sentiment analysis systems \cite{Xu2019}, and QA systems \cite{Huang2021,li2022question}.  In contrast to these existing works, LMPM is specifically designed to memorize and store the features of logical patterns, which can be reused and guide the model in generating intermediate conclusions of logical consistency.

\section{Methodology}
Formally, the entailment tree generation involves a hypothesis $Y$ and a set of factual sentences $S = \{ {s_1},{s_2}, \cdots ,{s_m}\} $ as inputs, where $m$ denotes the number of candidate facts. The hypothesis $Y$ can be deduced from a subset of $S$ and the goal is to generate a valid tree $T$ that portrays the deduction process, as depicted in Figure~\ref{FIG:1}. The leaves of the tree correspond to the facts selected from $S$, while the root refers to $Y$. We use $c_i$ to denote the intermediate node, representing the intermediate conclusion generated by the model. 
%Within the tree $T$, each intermediate node $I$ corresponds to a new fact (e.g., $I_1$ and $I_2$), while each non-leaf node $u \in \{ H,I\} $ corresponds to the conclusion and its children as the premises. 
Moreover, each non-root node, including both the leaves and intermediate nodes, also corresponds to a premise of its parent node.
% , indicated by an additional notation $p_i$\footnote{For example, $p_1$ in Figure~\ref{FIG:2}, which will be discussed later.}
%Each non-leaf node (i.e., the root or the intermediate node) corresponds to a reasoning step with the node as the conclusion and its children as premises.
% Therefore, the model needs to select relevant sentences from $S$ as the leaves, and uses them to compose a valid proof tree leading to the root $Y$, and fills the intermediate nodes using the generated intermediate conclusions.
% Based on the distractor settings of candidate facts $S$, 
Based on the scope of the candidate facts $S$,
the task can be further categorized into the following sub-tasks of increasing difficulty:
\\
 \textbf{Task 1} (no-distractor) $S = {S_{gold}}$ , \\
 \textbf{Task 2} (distractor) $S = {S_{gold}\ +\ }$15-20 distractors, \\
 \textbf{Task 3} (full-corpus) $S =$ full corpus. \\
where $S_{gold}$ is the set of golden leaf facts.

%\subsection{Logical Pattern Memory Pre-trained Model}
\subsection{LMPM}
In this section, we present the proposed \textbf{L}ogical Pattern \textbf{M}emory \textbf{P}re-trained \textbf{M}odel (LMPM), which consists of an encoder-decoder generative model (T5\footnote{We choose T5 as the backbone in LMPM, similar to the works previously introduced \cite{Dalvi2021, Yang2022, Hong2022}, to facilitate ease of comparison in experiments. Note that any encoder-decoder model can serve as the backbone, not exclusively T5.}) and an external memory denoted as \emph{M}. The overall architecture is illustrated in Figure~\ref{FIG:3}. 
The key insight of LMPM is to map the implicit logical patterns onto the embedding space. The training data of logical patterns, acquired through the entity abstraction techniques, force the model to concentrate specifically on essential logical connections, while simultaneously disregarding specific terms or domain-specific knowledge. 
For instance, consider the logical pattern depicted in the gray box of Figure~\ref{FIG:2}, which explores the concept of stages related to entity <E2>. The premises assert that both <E1> and <E3> represent stages of <E2>. From these premises, we can deduce the conclusion that <E1> and <E3> are distinct stages of <E2>. Thus, by leveraging such logical patterns, we can generalize the reasoning process and identify similar relationships across various instances.

%For example, when analyzing the logical pattern depicted in the gray box of Figure~\ref{FIG:2}, we can understand that it explores the concept of stages related to entity <E2>. The premises state that both <E1> and <E3> are stages of <E2>. From these premises, we can deduce the conclusion that <E1> and <E3> represent distinct stages of the <E2>. Thus, by leveraging the logical patterns, we can generalize the reasoning process and identify similar relationships across various instances. 
%The logical patterns refer to the crucial information that retains logical relationships (\emph{is a stage of} and \emph{are different stages of}) by removing the domain-specific terms (\emph{primitive society}, \emph{social development} and \emph{adolescence}).  These patterns encompass the fundamental structural elements that capture the underlying connections within a specific context. By directing our attention towards these patterns, we can extract and learn the core logical relationships among facts.

The memory \emph{M} is responsible for learning and storing representations of logical patterns between premises and conclusions. The T5 model takes the two premises as inputs, selects the logical pattern representations from \emph{M}, and leverages these representations to generate logically consistent conclusions during the deductive process. 
LMPM is trained through two essential tasks: \emph{Logical Pattern Pre-training} and \emph{EntailmentBank Fine-tuning}. The first task forces the model to capture and learn representations of latent logical patterns via a constructed entity-abstract dataset. 
The second task focuses on training the model to generate intermediate conclusions of logical consistency by adapting the pre-trained representations of logical patterns to the target EntailmentBank dataset.

%to adapt the pre-trained logical pattern representations to the target EntailmentBank dataset, and used the pattern representations to generate the intermediate conclusions of logical consistency.

%Subsequently, the $f_z$ component samples logical patterns from memory \emph{M} as the logical feature, and generates intermediate facts using additional logical features from the logical feature. LMPM is trained via two similar processes: \emph{logical Pattern Pre-training} and \emph{EntailmentBank fine-turning}. To enhance the ability of memory structure to capture latent logic patterns, we propose the Wiki data with entity abstraction (named logical pattern dataset) to train our proposed LMPM. The LMPM is then fine-tuned using EntailmentBank dataset to map the trained logical pattern from logical pattern dataset to the target dataset. During inference, the logical pattern is obtained from memory \emph{M} and enhances logicality in deductive generation with additional logical information.

\subsubsection{Logical Pattern Pre-training} 
As previously mentioned, the first task of LMPM aims to enhance its ability to capture and learn latent logical patterns. To achieve this, we construct a dataset using entity abstraction techniques. We utilize the Wikipedia-based synthetic dataset introduced by \citet{Hong2022}, which encompasses logical regularities but also contains a substantial amount of domain-specific information. Each data instance comprises two premises and a conclusion, simulating the process of generating intermediate conclusions. The logical deductive types are categorized as \emph{substitution}, \emph{conjunction}, and \emph{if-then}. Examples of logical patterns belonging to the conjunction type are illustrated in the green boxes of Figure~\ref{FIG:2}. 
To abstract the original synthesis data, 120 special tokens (e.g., ``<extra\_id\_1>" that used in T5) are leveraged to replace the entity spans. Specifically, we replace entity spans whose part-of-speech corresponds to \emph{noun}, \emph{det noun}, \emph{adj noun}, or \emph{def adj noun}. For instance, in Figure~\ref{FIG:2}, the phrase ``primitive society" within the top-left box is replaced with ``<E1>", and the gray box displays the result after the abstraction steps.
\iffalse
The resulting dataset, denoted as $N=\{(\tilde{ p}_1^{(n)},\tilde{p}_2^{(n)}),\tilde{c}^{n}\}_{n=1}^{N}$, consists of a collection of logical patterns. Here, $( \tilde{ p}_1^{(n)}, \tilde{p}_2^{(n)})$ and ${\Tilde c}^{(n)}$ represent the two pseudo premises and the corresponding pseudo intermediate conclusions, respectively.
The resulting dataset, denoted as $N=\{(\tilde{ p}_1^{(n)},\tilde{p}_2^{(n)}),\tilde{c}^{n}\}_{n=1}^{N}$, consists of a collection of logical patterns. Here, $( \tilde{ p}_1^{(n)}, \tilde{p}_2^{(n)})$ and ${\Tilde c}^{(n)}$ represent the two pseudo premises and the corresponding pseudo intermediate conclusions, respectively.
This constructed entity-abstract is utilized to learn a latent variable $f_z$, that is, the representation of a specific logical pattern stored in memory $M$, defined as: 
\begin{eqnarray}\label{eq:1}
\begin{array}{l}
{p({{\tilde c}^{(n)}},{f_z}|\tilde p_1^{(n)},\tilde p_2^{(n)})}\\
{ = p({f_z}|\tilde p_1^{(n)},\tilde p_2^{(n)})p({{\tilde c}^{(n)}}|{f_z},\tilde p_1^{(n)},\tilde p_2^{(n)})}
\end{array}
\end{eqnarray} 
\fi
The resulting dataset comprises a collection of logical patterns, where each data instance, denoted as $\{(\tilde{ p}_1, \tilde{p}_2), \tilde{c}\}$, consists of two pseudo premises and the corresponding pseudo intermediate conclusion. 
This entity-abstract dataset is constructed to learn a latent variable $f_z$ for each logical pattern, which is then utilized for generating logically consistent conclusions, and the equation is:
\begin{eqnarray}\label{eq:1}
\begin{array}{l}
{p({{\tilde c}},{f_z}|\tilde p_1,\tilde p_2)  = p({f_z}|\tilde p_1,\tilde p_2)p({{\tilde c}}|{f_z},\tilde p_1,\tilde p_2)}
\end{array}
\end{eqnarray} 

Based on the T5 encoder, we adopt a specific sequence format where we insert the token \emph{<s>} before the first premise, and use the token \emph{</s>} to either concatenate the two premises or indicate the sequence's end. Additionally, we introduce a special latent token \emph{<z>} at the beginning of the sequence. Thus, the input sequence $I_{{\mathop{\rm logical}\nolimits}}$ for logical pattern pre-training is transformed as follows:
\begin{eqnarray}\label{eq:2}
\small
\hspace{-2mm}
\begin{array}{l}
{I_{{\rm{logical}}}} = [ <\!\!z\!\!>,\!\!<\!\!s\!\!>,\tilde p_1^1,..,\tilde p_1^l, \!\! <\!\!/s \!\!> , \tilde p_2^1,...,\tilde p_2^{l'},\!\! <\!\!/s\!\!> ]
\end{array}
\end{eqnarray} 
where $l$ and ${l}'$ are the lengths of $\tilde p_1$ and $\tilde p_2$, respectively. Subsequently, we use the T5-encoder to encode the input sequence ${I_{{\mathop{\rm logical}\nolimits} }}$. The final hidden state of the token \emph{<z>}, denoted as $H_z$, serves as the logical representation of the two premises. 
$H_z$ is utilized to select the specific logical pattern from $M$. The external memory $M$, parameterized by $\theta$, is responsible for learning and storing the features of logical patterns. Each element $m_i$ in $M$ indicates a specific logical pattern:
\begin{eqnarray}\label{eq:3}
M = [{m_1},...,{m_L}] \in{\mathbb{R}} {^{L \times {d_m}}}
\end{eqnarray} 
where $L$ corresponds to the number of logical patterns stored in the memory, and $d_m$ is the dimension of the latent representations.
 \begin{figure}    
  \centering
    \includegraphics[scale=0.47]{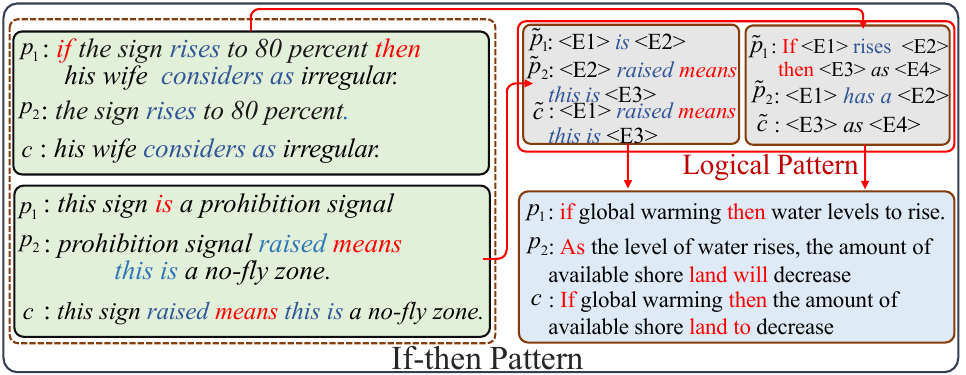}
    \vspace{-0.6cm}
  \caption{Examples of complex deductions. The gray boxes represent two distinct logical patterns that need to be combined to generate the appropriate conclusion shown in the blue box.}%Examples of the complex deductions. The two logical patterns depicted in the gray boxes need to be combined in order to generate the appropriate intermediate conclusion shown in the blue box.}
    \vspace{-0.2cm}
  \label{FIG:complex}
\end{figure}

%Given a training instance from the constructed entity-abstract dataset, which corresponds to a specific logical pattern, we aim to utilize its corresponding logical pattern representation for generation. Therefore, LMPM should be capable of selecting the potential logical pattern representations in $M$ and learning and optimizing the representations. 

To leverage the logical pattern representations stored in the memory $M$ for generating intermediate conclusions, LMPM must identify the specific logical pattern corresponding to the premises. This requires determining the value of $i$ that satisfies $f_z = m_i$, where $f_z$ denotes the representation of the target logical pattern, and $m_i$ refers to an element within the memory, as previously mentioned.
To facilitate this identification process and learn the representations, we introduce an address structure $D$, which employs a multi-layer perceptron architecture. This structure takes $H_z$ as input and produces a one-hot vector ${\alpha _i\in {\mathbb{R}}{^{1 \times L}}}$ to help locate the potential logical pattern representation in $M$. In particular, we use Gumbel-softmax \cite{jang2016categorical} for optimization:
\begin{eqnarray}\label{eq:4}
\begin{array}{c}
{\gamma _i} = {w_z}{H_z} + {b_z}\\
\end{array}
\end{eqnarray} 
\begin{eqnarray}\label{eq:42}
\begin{array}{c}
{\alpha _i} = \frac{{\exp ({\gamma _i} + {g_i})/{\cal T}}}{{\sum\limits_{i = 1}^L {\exp ({\gamma _i} + {g_i})/{\cal T}} }}
\end{array}
\end{eqnarray} 
Here, $g_i$ is a random variable following the Gumbel distribution and  $\cal T$ is the temperature coefficient. After obtaining the one-hot vector ${\alpha _i}$, we can locate the potential logical pattern in $M$ using:
\begin{eqnarray}\label{eq:5}
{f_z} = \sum\limits_{i = 1}^L {{\alpha _i}{M_i}}
\end{eqnarray} 

%Next, we integrate the selected logical pattern representations into the T5-decoder by performing an addition operation on the representation of the start special token <s>, with the object of generating an intermediate conclusion of logical consistency:
Next, we integrate the selected logical pattern representation into the T5-decoder by adding it to the representation of the start special token <s>. The purpose is to effectively introduce the logical pattern, as shown in the equation:
\begin{eqnarray}\label{eq:6}
E_{ <s > }^{{}'} = {f_z} + {E_{ <s > }}
\end{eqnarray} 

%The latent memory structure $M$ can provide the desired logical pattern representation by sampling the element. 
% Thus, the memory structure $M$ as well as the address structure $D$ can together be updated via backpropagation. Note also that the entity-abstract dataset of logical patterns is used for training in this task, and in Section 4.6, we show that such abstraction approach benefits the model not only to be less affected by the irrelevant knowledge, but also to be well trained with significantly less data.
%Thus, the memory structure $M$ and the address structure $D$ can be updated via backpropagation.
% Note also that the entity-abstract dataset of logical patterns is used for training in this task.
The memory structure $M$ and the address structure $D$ can be updated through backpropagation, utilizing the entity-abstract dataset. Additionally, Section 4.3 demonstrates that this abstraction approach offers substantial benefits to the model, notably reducing the impact of irrelevant knowledge and facilitating effective training with less data.

\noindent \textbf{Loss Function} 
To optimize the model parameters in the logical pattern pre-training task, two loss functions are employed. The first one is the language modeling loss (LM), which is defined as:
\begin{eqnarray}\label{eq:7}
{{\cal L}_{LM}} =  - \sum\limits_{t = 1}^{|\tilde{c}|} {\log p({\tilde{c}^{t}}|f_z,{\tilde{p}_1},{\tilde{p}_2},{\tilde{c}^{0:t-1}})} 
\end{eqnarray} 
where $|\tilde{c}|$ is the length of $\tilde c$. In addition, to tackle the issue of vanishing latent variables \cite{Zhao2017}, a bag-of-word loss (BOW) is applied: 
\begin{eqnarray}\label{eq:8}
\small
\begin{array}{c}
{L_{BOW}} = \sum\limits_{t = 1}^{|\tilde c|} {\log p({{\tilde c}^t}|{f_z},{{\tilde p}_1},{{\tilde p}_2})}  = \sum\limits_{t = 1}^{|\tilde c|} {\log \frac{{{e^{f({{\tilde c}^t})}}}}{{\sum\limits_{v \in V} {{e^{f(v)}}} }}} 
\end{array}
\end{eqnarray} 

Therefore, the total loss in the logical pattern pre-training is defined as follows:
\begin{eqnarray}\label{eq:10}
{{\cal L}} =  {{\cal L}_{LM}}+{{\cal L}_{BOW}}
\end{eqnarray}

\subsubsection{EntailmentBank Fine-tuning} 
The second task aims to fine-tune LMPM for generating actual intermediate conclusions with logical consistency in the entailment steps, adapting the pre-trained logical pattern representations to the target EntailmentBank dataset. 
While the overall process resembles the previous logical pattern pre-training task, it uses the actual premises and intermediate conclusions from the EntailmentBank dataset for training. Moreover, the dataset includes some complex deductions that can not be adequately captured by a single logical pattern stored in $M$. To address this limitation, the model must incorporate multiple logical pattern representations to generate appropriate intermediate conclusions, as illustrated in the example depicted in Figure \ref{FIG:complex}. To enable the combination of multiple logical pattern representations, we replace the Gumbel-softmax in equation (\ref{eq:42}) with Softmax:
% that the updated equation is: 
% The purpose of the second task is to train LMPM to generate the real intermediate conclusions of logical consistency in the entailment steps, by adapting the pre-trained logical pattern representations to the target EntailmentBank dataset. The overall process of this task is similar to the previous logical pattern pre-training task, but takes the real premises and intermediate conclusions of the entailment trees in the EntailmentBank dataset for training. However, EntailmentBank dataset contains some complex deductions that can not be effectively obtained by using single logical pattern stored in $M$. Multiple logical patterns are required to combined together to generate the appropriate intermediate conclusions, such as the example shown in the blue box in Figure \ref{FIG:3} (b). To address this limitation, we use Softmax to replace the Gumbel-softmax of eq (\ref{eq:42}), which is updated as: 
%EntailmentBank fine-tuning aims to map the logical pattern to the EntailmentBank dataset, which collect from the science domain. The overall framework of EntailmentBank fine-tuning is similar to logical pattern training. However, some complex deductions can not effectively map from single synthetic logical patterns, such as Figure \ref{FIG:3} (b). Therefore, we use Softmax to replace eq (\ref{eq:5}), which is defined as 
\begin{eqnarray}\label{eq:11}
\begin{array}{c}
\widetilde {{\alpha _i}} = \frac{{\exp ({\gamma _i})}}{{\sum\limits_{i = 1}^L {\exp ({\gamma _i})} }}
\end{array}
\end{eqnarray}
In this case, $\widetilde {{\alpha _i}}$ is not a one-hot vector. The model is trained to combine multiple logical pattern representations in $M$ to generate the conclusions.

\begin{table}[]
\centering
\begin{tabular}{c|ccc|c}
\toprule
                 & Train & Dev & Test  & All   \\
\hline

Entailment trees  & 1,131 & 187 & 340   & 1,840 \\

Entailment steps & 4,175 & 597 & 1,109 & 5,881\\
\bottomrule
\end{tabular}
% \vspace{-0.15cm}
\caption{EntailmentBank statistics.}
\label{tab:1}
% \vspace{-0.2cm}
\end{table}

\begin{table*}
\small
\centering
\setlength{\tabcolsep}{1mm}
% \usepackage{multirow}
% \usepackage{multirow}
\begin{tabular}{ccccccccc}

\toprule
 \multirow{2}{*}{Method}                 &Additional                             & \multicolumn{2}{c}{Leaves} & \multicolumn{2}{c}{Steps} & \multicolumn{2}{c}{Intermediates} & Overall    \\ 
 \cline{2-9}
 &Data                                                                  & $F_1$        & AllCorrect     & $F_1$        & AllCorrect    & $F_1$            & AllCorrect        & AllCorrect \\ 
\toprule

\multicolumn{9}{c}{Task1} \\
    \toprule

EntailmentWriter(T5-large)   &No                                       & 98.4     & 84.1           & 50.0     & 38.5          & 67.0          & 35.9              & 34.4         \\ 

 % & EntailmentWriter(T5-11B)   &No                                       & 99.0     & 89.4           & 51.5     & 38.2          & \underline{71.2}          & \textbf{52.9}              & 35.6        \\ 
 IRGR &No                                       & 97.6     &89.4 &50.2 &36.8 &62.1 &31.8 &32.4        \\ 
 RLET  &No                                       & \textbf{100.0} &\textbf{100.0}  &55.0 &41.2 &67.2 &36.7& 35.1        \\

% \cline{2-10}
 METGEN+T5         &564K                                                  & \textbf{100.0}       & \textbf{100.0}            & \underline{57.7}      & \underline{41.9}          & \underline{70.8}          & \underline{39.2}              & \underline{36.5}       \\ \

\begin{tabular}[c]{@{}c@{}}\textbf{METGEN+LMPM(\emph{ours})}\end{tabular}  &564K & \underline{99.76}    & \underline{99.41}        & \textbf{57.78}     & \textbf{43.82}{\dag}         & \textbf{72.78}{\dag}         & \textbf{42.78}{\dag}             & \textbf{38.54}{\dag}      \\ 
\toprule
\multicolumn{9}{c}{Task2} \\
    \hline
% \cline{2-10}
                       % & NLProofS+T5            &No                                               & 98.7      & 94             & 56.6      & \underline{42.9}          & \underline{74.4}          & 42.5              & \underline{39.9}       \\ \

                       % & \begin{tabular}[c]{@{}c@{}}\textbf{Nlproofs+LMPM(\emph{ours})}\end{tabular} &564K & 97.7      & 92.47          & 54.75     & 42.35         & \textbf{74.75}{\dag}         & \textbf{42.69}{\dag}             & \textbf{40.29}{\dag}     
                       % \\ 
                       \hline
EntailmentWriter(T5-large)       &No                                       & \textbf{83.2}      & 35.0           & 39.5    & 24.7         & \textbf{62.2}          & 28.2             & 23.2        \\  
% \cline{2-10}

 % & EntailmentWriter(T5-11B)   &No    &\textbf{89.1} &\textbf{48.8} &\underline{41.4} &27.7 & \textbf{66.2} &\textbf{53.2} &25.6        \\ 
 
IRGR &No                                       & 69.9 &23.8 &30.5 &22.40 &47.70 &26.5 &21.8        \\ 
RLET  &No                                       &81.9 &40.4 &38.8 &28.7 &57.4 &29.1 &26.0
        \\

 METGEN+T5    &564K                                                       & \underline{82.7}      & \underline{46.1}           & \underline{41.3}      & \underline{29.6}          & 61.4          & \underline{32.4}              & \underline{27.7}       \\ \
 \begin{tabular}[c]{@{}c@{}}\textbf{METGEN+LMPM(\emph{ours})}\end{tabular} &564K   & 81.09     & \textbf{47.06} 
                               & \textbf{42.56}     & \textbf{31.38} {\dag}         & \underline{61.68}        & \textbf{34.32}{\dag}             & \textbf{29.41}{\dag}      \\ 
\toprule
\multicolumn{9}{c}{Task3} \\
    \hline

                      %   & NLProofS+T5      &No                                                     & \textbf{90.3}      & \textbf{58.8}           & \textbf{47.2}      & 34.4          & \textbf{70.2}          & \textbf{37.8}              & \underline{33.3}       \\ I am running a few minutes late; my previous meeting is running over. 
                      %  & \begin{tabular}[c]{@{}c@{}}\textbf{NLProofS+LMPM(\emph{ours})}\end{tabular} &564K & \underline{88.97}     & \underline{50.29}          & \underline{46.11}     & \textbf{35.59}         & \underline{67.05}{\dag}         & \underline{37.67}{\dag}             & \textbf{33.82}{\dag}      \\ 
                      \hline
EntailmentWriter(T5-large)   &No                                            & 30.9    & 1.2           & 4.4       & 1.2           & 28.8          & 5.6              & 1.2        \\ 
% \cline{2-10}

 % & EntailmentWriter(T5-11B)   &No    &39.7 &3.8 &7.8 &2.9 &36.4 &13.2 &2.9        \\ 
 
 IRGR &No                                       & \textbf{46.6} &\underline{10.0} &\underline{11.3} &8.2 &\underline{38.7} & \textbf{20.9} &8.2        \\ 
 RLET  &No                                       & \underline{46.20} &\textbf{11.41} &\textbf{15.2} &\textbf{9.6} &\textbf{41.4} &17.6 & \underline{9.4} \\

METGEN+T5      &564K                                                     & 34.8      & 8.7            & 9.8       & 8.6           & 36.6          & \underline{20.4}              & 8.6        \\ 
\begin{tabular}[c]{@{}c@{}}\textbf{METGEN+LMPM(\emph{ours})}\end{tabular}&564K   & 35.3      & 9.23           & 10.28     & \underline{9.23}          & 37.8          &20.33            & \textbf{9.41}       \\ 
% \cline{2-10}
                   % & NLProofS+T5       &No                                                  & \textbf{43.2}      & 8.2            & \textbf{11.2}      & 6.9           & \textbf{42.9}          & 17.3              & 6.9        \\ \
                   %     & \begin{tabular}[c]{@{}c@{}}\textbf{NLProofS+LMPM(\emph{ours})}\end{tabular}&564K & 32.19     & 8.82           & 9.85      & 7.05          & 34            & 17.05             & 6.76       \\ 
\bottomrule

\end{tabular}
\vspace{-0.3cm}
\caption{Automatic evaluation results on the EntailmentBank dataset (\%). The best and second-best results are highlighted in bold and underlined, respectively. {\dag} refers to significance test $p-value<0.05$. The ``Additional Data" column denotes the size of the supplementary Wikipedia training data.}
\label{table:2}
\vspace{-0.5cm}
\end{table*}

\section{Experiments}
\subsection{Datasets \& Implementation Details}

We perform experiments on EntailmentBank \citep{Dalvi2021}, which is created based on ARC dataset of grade-school science questions \citep{clark2018think}, and a corpus of science and general knowledge from WorldTree V2 \citep{Xie2020}. EntailmentBank contains 1840 entailment trees, where each tree corresponds to a question. The statistics are shown in Table \ref{tab:1}.

% EntailmentBank contains 1840 entailment trees, where each tree corresponds to a question. 
%\subsection{Implementation Details}
% \subsection{Implementation Details}

The memory structure stores $L=7$ logical patterns, corresponding to the number of inference types in the EntailmentBank dataset. Each memory element has a dimension of $d_m$ set to 4096. During the logical pattern pre-training phase, the model is trained for two epochs using a batch size of 35 and a learning rate of 3e-5. For the subsequent EntailmentBank fine-tuning phase, the model is trained for 80 epochs with a batch size of 30 and a learning rate of 3e-5.
To create the logical pattern pre-training dataset, we employ spaCy\footnote{https://spacy.io/models.} with the \emph{en\_core\_web\_sm} version to parse sentences and extract the part-of-speech information for each token. Subsequently, we use GloVe \cite{JeffreyPenningtonRichardSocher2014} to identify tokens of different forms in the premises and conclusions, replacing them with the same special token. Regarding the quantity and length of logical patterns, our constructed dataset comprises a total of 565,453 patterns, with an average length of 17 and a maximum length of 43. %It should be noted that these lengths fall within the confines of T5's limit of 512.
 Notably, in the experiments, we integrate LMPM with METGEN by replacing its T5 component, allowing it to utilize the intermediate conclusions of logical consistency.

\subsection{Evaluation Metrics}
\textbf{Automatic Evaluation} We assess the generated tree in comparison to the golden tree across three dimensions, following established practices in previous studies \citep{Dalvi2021, Hong2022, Yang2022}: 
(1) \textbf{Leaves:} measures whether the correct leaf facts are used. We compute the $F_1$ score by comparing the predicted leaf facts to those in the golden tree. Additionally, we report the \textit{AllCorrect} score, which indicates exact matches\footnote{The AllCorrect is 1 if the $F_1$ is 1, and 0 otherwise.}.
(2) \textbf{Steps:} evaluates whether the entailment steps are structurally correct. We compare the steps in two trees using the $F_1$ score and \textit{AllCorrect}. A predicted step is considered correct if its children's identifiers perfectly match the corresponding ones in the gold tree.
(3) \textbf{Intermediates:} assesses the accuracy of intermediate conclusions, also utilizing the $F_1$ score and the \textit{AllCorrect} score. A predicted intermediate conclusion is considered correct if the \texttt{BLEURT-Large-512} \cite{sellam2020bleurt} score between the aligned predicted intermediate and the corresponding golden intermediate is greater than 0.28.
% In addition to these metrics, we employ the \textit{Overall AllCorrect} measure, which equals 1 only if all the leaves, steps, and intermediates are correct according to their \textit{AllCorrect} criteria. It represents the strictest evaluation measure.
% Note that in automatic evaluation, we test the models on three tasks based on different distractor settings in EntailmentBank, as introduced in Section 3.

\textbf{Human Evaluation} 
% To address potential limitations of automatic evaluation, as highlighted by \citet{Dalvi2021}, which may misjudge some valid trees and underestimate performance, we conduct human evaluation to compare the performance of different methods.
Considering the potential limitations of automated evaluation, as highlighted by \citet{Dalvi2021}, which might inaccurately assess certain valid trees and underestimate their performance, we conducted a human evaluation. We randomly sampled 100 instances from the EntailmentBank test set and asked 9 graduate students to evaluate the model results based on four criteria:
%We also conduct human evaluation to compare the quality of the generated results between our method and the baselines, using four criteria: 
%For a more comprehensive comparison of our model with baselines, we set the four manual metrics to respectively evaluate the overall and generated intermediate conclusions: 
(1) \textbf{Validity}: evaluates whether the generated tree clearly demonstrates a coherent chain of reasoning, explaining how the hypothesis is derived from the supporting facts. 
(2) \textbf{Logical consistency}: assesses whether each intermediate conclusion aligns with the given premises. 
(3) \textbf{Readability}: assesses whether the generated intermediate conclusions are grammatically correct and fluent to read.
(4) \textbf{Reasonability}: evaluates whether the generated intermediate conclusions adhere to common knowledge and facts. 
%The first metric is based on the entire entailment trees and we use the opinions of the majority to resolve disagreements on some generated trees results. Thus, the metric 
For the \textbf{Validity} metric, we compute the number of trees considered valid by the majority of evaluators. For the remaining three criteria, the students were asked to rate the generated intermediate conclusions from 1 (very bad) to 5 (very good), and the average scores were reported. Note that the human evaluation was only conducted for task 1, as the other two tasks would have been more challenging and time-consuming for human evaluators.
%For the \textbf{Validity} metric, we computed the number of trees considered valid by the majority of evaluators. For the remaining three criteria, the students were asked to rate the generated intermediate conclusions on a scale from 1 (very bad) to 5 (very good), and the average scores were reported. Notably, the human evaluation was only conducted for Task 1, as the other two tasks would have been more challenging and time-consuming for the human evaluators.

% To obtain the logical pattern pre-training dataset, we apply the spaCy\footnote{https://spacy.io/models.} with \emph{en\_core\_web\_sm} version to parse the sentences and obtain the part-of-speech for each token. Then we use GloVe \cite{JeffreyPenningtonRichardSocher2014} to find the token of different forms in the premises and conclusions, that will be replaced by the same special token. $L$, the number of logical patterns stored in the memory structure, is set to 7 equaling to the number of inference types in EntailmentBank. 
% And the element dimension $d_m$, i.e., the dimension of the logical pattern representation, is set to 2048. 
% In the logical pattern pre-training phase, we set batch size as 35, learning rate as 3e-5 and train the model for two epochs. In the EntailmentBank fine-tuning phase, we set batch size as 30, learning rate as 3e-5 and train for 80 epochs.
%\footnote{We implement our model with the PyTorch framework. and conduct experiments on the machine with NVIDIA Quadro RTX 8000.}

\begin{table}[]
\small
\centering
\begin{tabular}{c|c|c}
\toprule
Mtrics  & METGEN+T5 & METGEN+LMPM \\
\hline
Validity       & 46        & 52          \\
Logical & 3.02      & 3.37        \\
Readability        & 4.11      & 4.43        \\
Reasonability      & 3.37      & 3.64       \\
\bottomrule
\end{tabular}
\vspace{-0.2cm}
\caption{Human evaluation results on 100 uniformly sampled questions from the test split.} %The automatic and manual, respectively, represent the correct percent (\%) of automatic and manual evaluation, and logical denotes the logical consistency.}

\label{table:3}
\vspace{-0.4cm}
\end{table}

%test some existing multi-step generative methods and evaluate the models incorporated with the LMPM model. 

\subsection{Experimental Results}
% \subsubsection{Datasets}
% \begin{table}[]
% \centering
% \begin{tabular}{c|ccc|c}
% \toprule
%                  & Train & Dev & Test  & All   \\
% \hline

% Entailment trees  & 1,131 & 187 & 340   & 1,840 \\

% Entailment steps & 4,175 & 597 & 1,109 & 5,881\\
% \bottomrule
% \end{tabular}
% % \vspace{-0.15cm}
% \caption{EntailmentBank statistics.}
% \label{tab:1}
% % \vspace{-0.2cm}
% \end{table}
% We conduct experiments on EntailmentBank \citep{Dalvi2021}, which is created based on ARC dataset of grade-school science questions \citep{clark2018think}, and a corpus of science and general knowledge derived from WorldTree V2 \citep{Xie2020}. EntailmentBank contains 1840 entailment trees, where each tree corresponds to a question. 
% The statistics are shown in Table \ref{tab:1}.
\subsubsection{Automatic Evaluation Results}
% The automatic evaluation results are summarized in Table~\ref{table:2}. We can see that the multi-step methods (METGEN, NLProofs) achieve better performance than the single-step method (EntailmentWriter) in most cases. Especially on the Overall AllCorrect metric, the best method outperforms EntailmentWriter by about 5.5\%, 10.1\% and 7.4\% in task 1, task 2, and task 3, respectively. This indicates that the multi-step method can more effectively select the premises and generate the intermediate conclusion more aligned with the ground truth.  
The automatic evaluation results are summarized in Table~\ref{table:2}. 
In general, the multi-step methods (IRGR, RlET, and METGEN) 
% In general, the multi-step methods (METGEN, NLProofs) 
outperform the single-step method (EntailmentWriter). This underscores the effectiveness of multi-step methods in selecting premises and generating intermediate conclusions that align closely with the ground truth. Benefiting from its endeavor to pre-train with additional Wikipedia-based synthetic data, METGEN has demonstrated robust and competitive performance, significantly outperforming the baseline models in both task 1 and task 2.

When comparing the performance of METGEN using different modules (T5 or LMPM) for intermediate conclusion generation, we observe a clear performance improvement when incorporating our proposed LMPM, especially in tasks 1 and task 2. Specifically, under the strictest \emph{Overall AllCorrect}, the model with LMPM achieved performance improvements of 2.04\% and 1.7\%, respectively. The enhancements under the \emph{Intermediates AllCorrect} metric further suggest that LMPM generates better intermediate conclusions with improved logical consistency. Additionally, the improved \emph{Steps AllCorrect} indicate that LMPM positively impacts the premise selection process of the controller in METGEN. Although the incorporation of LMPM within METGEN generally outperforms the use of T5 in task 3, the improvements are not as pronounced. This could be attributed to the fact that task 3 utilizes the full-corpus distractor setting, which heavily relies on the accurate selection of premises.
% Furthermore, the NLProofs model, when utilizing LMPM for conclusion generation, also demonstrates improved results under the \emph{Overall AllCorrect} metric for tasks 1 and 2. This finding underscores the potential applicability of our model to verification-based methods. Thus, LMPM can offer effective logical pattern features that contribute to the generation of more accurate conclusions in entailment steps, thereby facilitating entailment tree generation.

% However, we also observe that the improvement on task 3 is not obvious as this task uses the full-corpus distractor setting, which relies heavily on selecting correct premises. Furthermore, the NLProofs model using LMPM for conclusion generation also achieves better results under Overall AllCorrect metric on task 1 and task 2, showing that our model could be applied to such verified-based methods. 
% Thus, LMPM can offer effective logical pattern features for generating better conclusions in the entailment steps, aiding in the entailment tree generation.

%By abstracting the Wiki data and using the external memory structure, the LMPM can effectively capture the logical pattern from the per-training step and use the logical pattern to guild the generation process of intermediate conclusions. 
%using LMPM to replace the vanilla T5 in NLProofs also achieved consistent performance improvement. The ability of vanilla T5 in NLProofs is not over overlapping our training objectives, showing that our model has applicability to verified-based methods. 

\begin{table}
\small
\centering
\setlength{\tabcolsep}{1mm}
% \usepackage{multirow}
% \usepackage{multirow}
\begin{tabular}{c|cc|cccc}
\toprule
\multirow{2}{*}{Method} & \multicolumn{2}{c|}{Task 1}  & \multicolumn{2}{c}{Task 2}  \\
                        & Inter & Overall    & Inter & Overall    \\
\hline
METGEN+T5                 & 39.2          & 36.5       & 32.4          & 27.7       \\
METGEN+LMPM             & 42.78         & 38.54      & 34.32         & 29.41      \\
\hline
w/o \emph{LPP}    & 39.02        & 36.47      & 32.24         & 27.56    \\
w/o \emph{memory}              & 40.59         & 37.39      & 33.06         & 28.54      \\
w/o \emph{abstraction}                & 41.27         & 37.45      & 33.78         & 28.48      \\
\bottomrule
\end{tabular}
\vspace{-0.25cm}
\caption{Ablation study results. ``Inter" and ``Overall" denote Intermediates AllCorrect and Overall AllCorrect, respectively.}
%The Inter and AC respectively denote the intermediates and AllCorrect}
\label{table:4}
\vspace{-0.4cm}
\end{table}

\begin{figure*} %这里使用的是强制位置，除非真的放不下，不然就是写在哪里图就放在哪里，不会乱动
  \centering  %图片全局居中
    % \small
  % \vspace{-0.35cm} %设置与上面正文的距离
  \subfigtopskip=2pt %设置子图与上面正文或别的内容的距离
  \subfigbottomskip=2pt %设置第二行子图与第一行子图的距离，即下面的头与上面的脚的距离
  \subfigcapskip=-5pt %设置子图与子标题之间的距离
  \subfigure[Task 1.]{
    \label{level.sub.1}
    \includegraphics[width=0.36\linewidth]{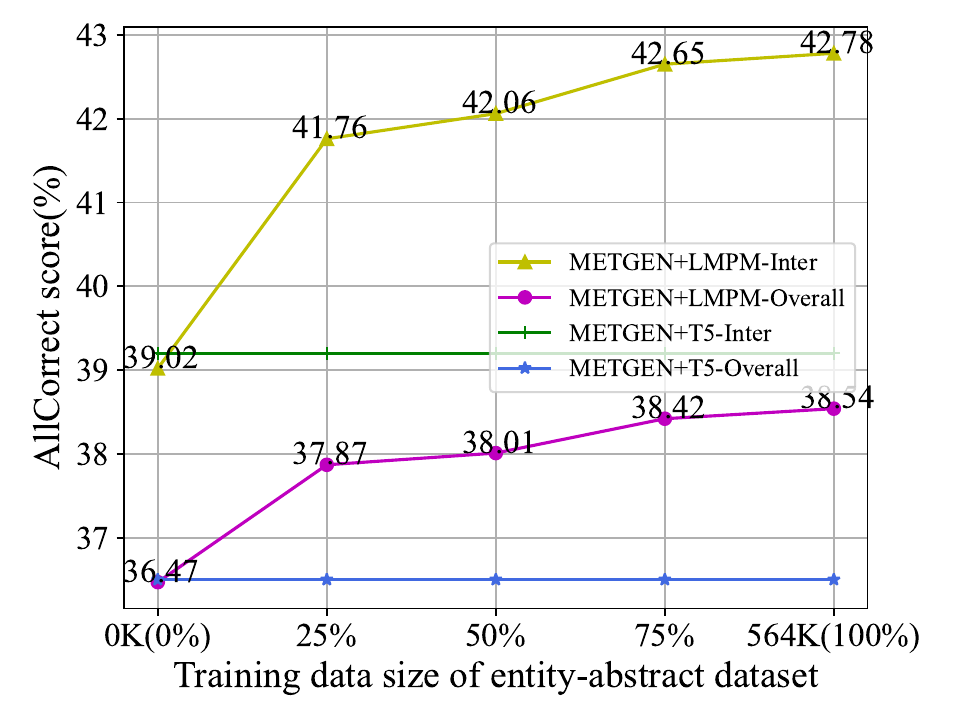}}
  \quad %默认情况下两个子图之间空的较少，使用这个命令加大宽度
  % \quad %默认情况下两个子图之间空的较少，使用这个命令加大宽度
  \quad %默认情况下两个子图之间空的较少，使用这个命令加大宽度
  \subfigure[Task 2.]{
    \label{level.sub.2}
    \includegraphics[width=0.36\linewidth]{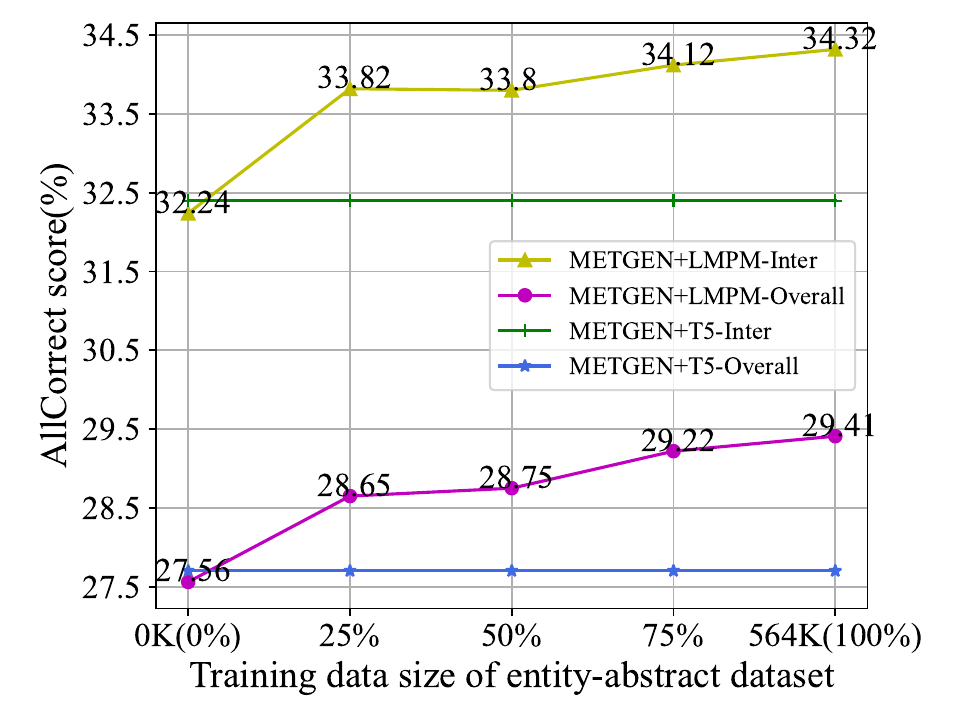}}
    \vspace{-0.3cm}
  \caption{The impact of logical pattern pre-training data size on performance. The AllCorrect scores for Overall and Inter (Intermediates) are provided.}
 %Inter and Overall denotes the Intermediate and overall, respectively.} %The number effect of logical pattern datasets. The inter denotes the Allcorrect score of the intermediates and T5 represents the T5 model used by METGEN, which is trained on 564K of Wiki synthetic datasets. }
  \label{Fig:4}
  \vspace{-0.3cm}
\end{figure*}

\begin{figure}    
  \centering
    \includegraphics[scale=0.3]{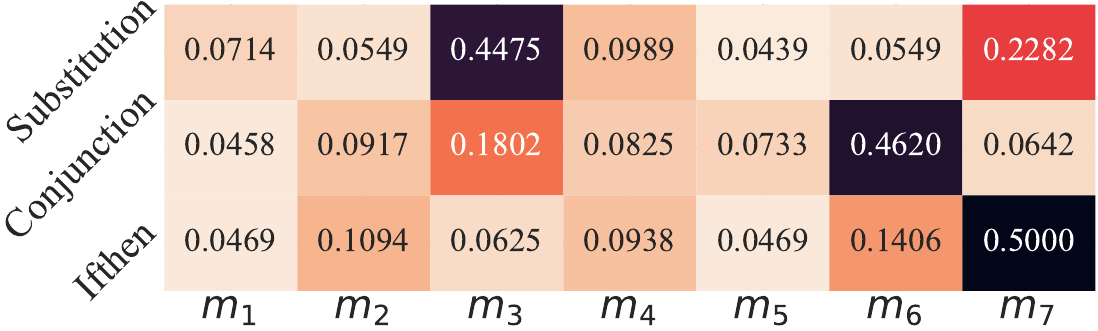}
    \vspace{-0.4cm}
  \caption{Probability distribution of the three logical patterns within the memory structure $M$.} %for the generation of intermediate conclusions using the select premises.}
  \label{FIG:attention}
\vspace{-0.4cm}
\end{figure}
\subsubsection{Human Evaluation}
%on the entailment trees generated by the original METGEN using T5 and the METGEN using LMPM. 
%We then conduct human evaluation to study the quality of the generated trees by different approaches.
In the human evaluation, we randomly sample 100 instances from the test set and compare the results generated by METGEN with different modules (T5 or LMPM). The results are presented in Table~\ref{table:3}. METGEN+LMPM exhibits superior performance compared to METGEN+T5 across all metrics, particularly in terms of \emph{Validity}, which indicates the quality of the complete entailment trees. In addition, \emph{Logical} and \emph{Readability} metrics also show enhanced performance, indicating LMPM's capacity to generate superior intermediate conclusions with logical consistency.

%METGEN+LMPM exhibits superior performance compared to METGEN+T5 across all metrics, particularly in terms of \textbf{Validity}, which serves as an indicator of the quality of the complete entailment trees. Furthermore, there is an evident improvement in the \textbf{Logical} and \textbf{Readability} aspects, suggesting that LMPM effectively generates intermediate conclusions that are logically consistent. Moreover, textbf{Logical} and \textbf{Readability} metrics also show enhanced performance, indicating that LMPM generates intermediate conclusions with better logical consistency.

% Moreover, Appendix~\ref{appendix_A_1} provides an analysis of three examples of generated entailment trees as a case study. % in the entailment steps. 
%Note that the criteria of Logical consistency, Readability, and Reasonability is to evaluate the quality of the generated intermediate conclusions whereas the Validity is to evaluate the entire entailment tree, as explained in Section 4.2.
%Thus, compared with T5, our proposed model LMPM can generate better conclusions of logical consistency, which could also benefit the models in obtaining better entailment tree. 

\subsubsection{Ablation Study}
In the ablation study, we delve into the effectiveness of three key components: \textit{logical pattern pre-training (LPP)}, \textit{memory structure (memory)}, and \textit{dataset abstraction (abstraction)}. Our analysis is conducted on both task 1 and task 2, with results presented in Table~\ref{table:4}.

Firstly, we examine the model without \textit{LPP}, which is exclusively trained based on the EntailmentBank Fine-tuning task. This model exhibits a significant performance degradation, highlighting the critical role of \textit{LPP} in generating reasonable intermediate conclusions. However, even without \textit{LPP}, the model's performance is on par with METGEN+T5, which is first pre-trained using the original synthetic dataset and then trained using the EntailmentBank. This suggets that the architecture of LMPM is more suitable than vanilla T5 in the entailment task.
Next, we compare the model without a \textit{memory structure}, having the same architecture as METGEN+T5 but first pre-trained on the entity-abstract dataset. The degraded performance of this model emphasizes the importance of the \textit{memory structure}, which stores and learns logical patterns. Despite this, we observe that this model still outperforms METGEN+T5, indicating that the pre-training abstraction data can help generate better conclusions. This is further proved by the model's decreased performance when trained without \textit{dataset abstraction} when incorporating LMPM. Finally, the model with the complete LMPM, which incorporates all the components simultaneously, achieves the best results.
\vspace{-0.3cm}
\subsection{Effect of Logical Pattern Data Size}
We next investigate the influence of data sizes in the pre-trained entity-abstract dataset on both task 1 and task 2. To achieve this, we employ METGEN+LMPM and train the model using various data sizes. We also present the results of METGEN+T5, which is pre-trained using the entire Wikipedia-based synthetic dataset. The results are shown in Figure~\ref{Fig:4}. Notably, a 0\% data size implies that the entity-abstract dataset is not utilized, and the model is exclusively trained via the EntailmentBank Fine-tuning process, aligning with the model without LPP as detailed in Section 4.5.3. Additionally, we observe that even with a mere 25\% of the data, the model achieves comparable performance to that using the entire dataset for both tasks. The entity-abstract dataset, in contrast to the original Wikipedia-based synthetic dataset, contains substantially less irrelevant knowledge within each logical pattern. This crucial distinction encourages the model to concentrate more precisely on the logical patterns, enabling effective training with a smaller amount of data. 

\vspace{-0.2cm}
\subsection{Additional Analysis of Logical Pattern Representation}
To gain a comprehensive understanding of the $L$ logical patterns stored in the memory structure ($L=7$ in the experiments), we further investigate the probability distribution of three different logical patterns within the memory structure $M$. We analyze a sample of 275 intermediate conclusions provided by METGEN, along with their automatically annotated inference types. The distribution of these samples is as follows: Substitution (153), Conjunction (71), and IF-then (51).

As depicted in Figure \ref{FIG:attention}, we observe distinct clustering tendencies for different logical patterns. Specifically, Substitution mainly correlate with logical feature $m_3$ within memory $M$, while Conjunction is closely associated with feature $m_6$, and IF-then primarily relates to feature $m_7$. These observed patterns highlight the model's capability to capture and differentiate various types of logical patterns. However, we also acknowledge that our model's relative neglect of a specific logical feature can be attributed to certain factors, such as potential errors introduced during the automated annotation process and instances necessitating a fusion of multiple logical pattern features.

\section{Conclusion}
In this paper, we present the logical pattern memory pre-trained model (LMPM), which significantly enhances the generation of intermediate conclusions in entailment steps by effectively harnessing logical patterns. LMPM utilizes an external memory structure to learn and store the latent representations of logical patterns, leveraging them to produce logically consistent conclusions. 
The model is pre-trained using a specially constructed entity-abstract dataset, tailored to facilitate the explicit acquisition of logical pattern features. Comprehensive evaluations demonstrate the superior ability of LMPM in facilitating the generation of better entailment trees, offering promising prospects in addressing the critical challenge of logical reasoning and paving the way for further explorations in complex reasoning tasks. For future work, we plan to extend this method to other tasks, such as the Multimodel QA, and to explore the incorporation of external knowledge to enhance tree generation.

\section*{Limitations}
While the LMPM model demonstrates promising capabilities, several limitations need to be acknowledged. 
Firstly, the lack of diverse datasets for entailment tree generation in languages other than English hampers the comprehensive evaluation of LMPM's performance across multiple languages. Consequently, the model's effectiveness in generating logically consistent conclusions is primarily demonstrated within English entailment tree corpora.
Secondly, LMPM may encounter challenges when tasked with generating logically consistent conclusions in scenarios involving more than two premises. Its performance in such cases might exhibit degradation, emphasizing the need for further enhancements to effectively manage complex entailment scenarios with multiple premises.

\section*{Acknowledgments}
This work was supported by the National Natural Science Foundation of China (62076100), Fundamental Research Funds for the Central Universities, SCUT (x2rjD2230080), the Science and Technology Planning Project of Guangdong Province (2020B0101100002), CAAI-Huawei MindSpore Open Fund, CCF-Zhipu AI Large Model Fund.

\nocite{*}
\section*{References}\label{sec:reference}

\bibliographystyle{lrec-coling2024-natbib}
\bibliography{lrec-coling2024-example}

\bibliographystylelanguageresource{lrec-coling2024-natbib}
\bibliographylanguageresource{languageresource}
\newpage
\appendix
\section{Appendix}

\subsection{Baselines}
To evaluate the effectiveness and applicability of our proposed method, we apply LMPM to several existing multi-step generative methods and compare the performance of the integrated model with the original methods.
The first multi-step method we compared is \textbf{METGEN} \citep{Hong2022}, which uses a reasoning controller and a T5 model to independently select premises and generate a conclusion. Note that the original synthesis Wikipedia dataset is first used to pre-train the METGEN model. In our comparison, we incorporate LMPM into METGEN by replacing their T5 model for intermediate conclusion generation.
The second compared multi-step method is \textbf{NlProofS} \citep{Yang2022}, which uses a T5 to simultaneously select the premise and generate the conclusion, followed by a verifier. In our comparison, we integrated LMPM into NlProofS by maintaining their T5 model for premise selection and incorporating LMPM to generate intermediate conclusions. 
%To apply LMPM in NlProofS, we keep their T5 to select the premises and use LMPM to generate the intermediate conclusions. 
% To verify the effectiveness and applicability of our proposed method, we apply LMPM to some existing multi-step generative methods and also compare with the original methods.
% The first multi-step method we compared is \textbf{METGEN} \citep{Hong2022}, which uses a reasoning controller and a T5 to select premises and generate a conclusion independently. Note that the original synthesis Wikipedia dataset is first used to pre-train the model. We can simply apply LMPM in METGEN by replacing their T5, which is for intermediate conclusion generation. 
% The second compared multi-step method is \textbf{NlProofS} \citep{Yang2022}, which uses a T5 to select the premise and generate the conclusion simultaneously, followed by a verifier. To apply LMPM in NlProofS, we keep their T5 to select the premises and use LMPM to generate the intermediate conclusions. 
Note also that the EntailmentBank dataset contains some entailment steps (9.1\%, 535 steps) where the intermediate conclusions are generated by more than two premises. METGEN decomposes such n-premise steps (n > 2) into several valid 2-premise steps before generating the conclusion. On the other hand, NLProofs can directly generate conclusions using more than two premises. Therefore, when applying LMPM to NLProofs, we retain their generated conclusions by T5 for the n-premise steps (n > 2). In all models mentioned, we freeze the parameters of the controller or verifier. Additionally, we use EntailmentWriter \citep{Dalvi2021} as a baseline, which generates the entire entailment trees in one step.

%As their T5 performs the two steps simultaneously, we cannot replace it directly. Thus, we keep the premise selection results by T5 and use LMPM to generate the conclusions when testing the application of LMPM to NlProofS.
%To further verify the applicability of our model, we apply LMPM in \textbf{NlProofS} \citep{Yang2022}, which uses vanilla T5 to select the premise and generated conclusion simultaneously, followed by a verifier to verify the steps. 
%The LMPM concentrate on generating a conclusion containing two premise steps. Thus, we keep the generated conclusion from NLProofs with more than two premises. 

\subsection{Case Study}
\label{appendix_A_1}
% In this part, we use three entailment trees generated by METGEN-T5 and METGEN-LMPM, that also the three trees used in human evaluation for the case study. For each tree, the task input, the golden tree, the trees generated by METGEN-T5 and METGEN-LMPM, and their corresponding Overall AllCorrect and human evaluation results\footnote{Note that Overall AllCorrect is the strictest automatic evaluation metric. And in Validity, we also show how many of the 9 evaluators consider the given generated tree is valid.} are presented, as shown in Figure~\ref{FIG:5}, ~\ref{FIG:6}, and  ~\ref{FIG:7}. We also show the types of pattern relations identified LMPM in the sub-graph of METGEN-LMPM.

\begin{figure*} %这里使用的是强制位置，除非真的放不下，不然就是写在哪里图就放在哪里，不会乱动
    \includegraphics[scale=0.6]{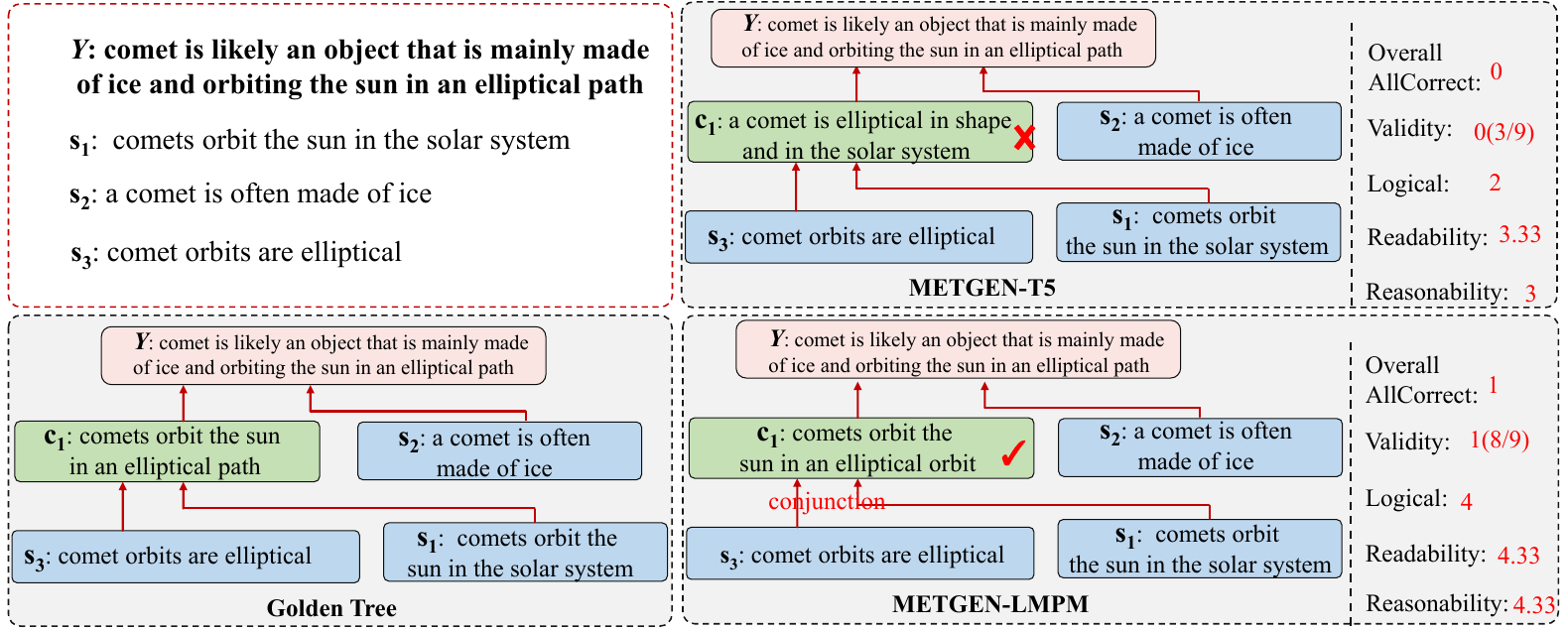}
    \caption{Case 1. A case shows that METGEN-LMPM achieves better results than METGEN-T5 in Overall AllCorrect metric and all metrics in human evaluations, and generates a better intermediate conclusion with logically consistency.}
  %\caption{Case 1. A case shows that the quality of intermediate conclusions effects the score of Overall AllCorrect.  The Intermediate AllCorrect score T5 is 0; the Overall AllCorrect score is 0 in Automatic evaluation. The proposed LMPM generated more similar with Gold intermediate conclusion and achieved Intermediate AllCorrect score of 1. Human denotes the percent of experts consider this is a correct predict tree.}
  \label{FIG:5}
\end{figure*}

\begin{figure*} %这里使用的是强制位置，除非真的放不下，不然就是写在哪里图就放在哪里，不会乱动
  \centering  %图片全局居中
    \includegraphics[scale=0.5]{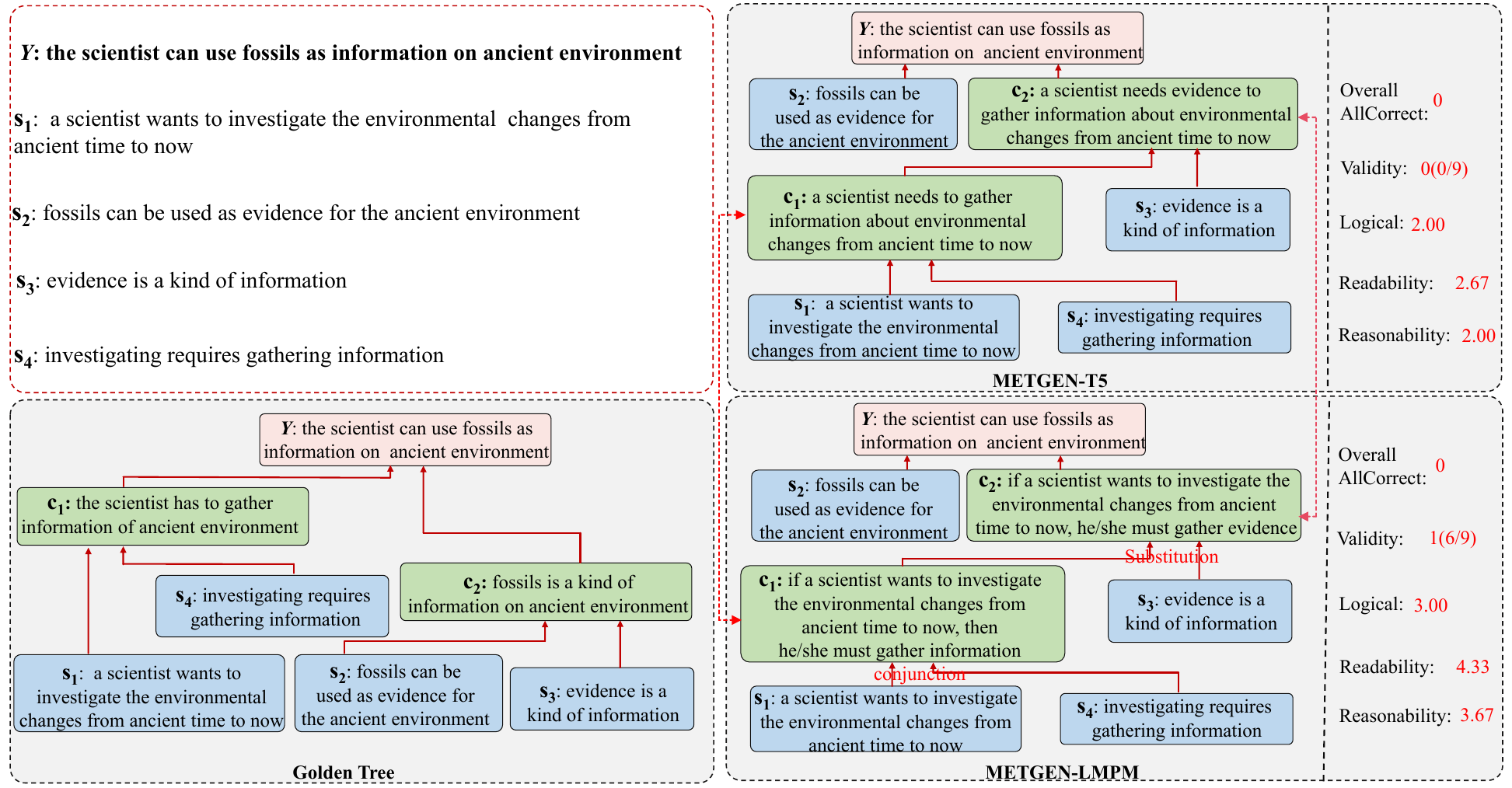}
    \caption{
Case 2: An interesting case shows that automatic evaluation can misjudge certain valid trees, leading to an underestimated performance. This observation is based on the results obtained from METGEN-LMPM, where the generated entailment tree is considered valid by the majority of evaluators.}
  %\caption{A case illustrates that LMPM can fuse the logical pattern between premises and generate more logical consistency and reasonability intermediate conclusions.}
  \label{FIG:6}
\end{figure*}

\begin{figure*} %这里使用的是强制位置，除非真的放不下，不然就是写在哪里图就放在哪里，不会乱动
  \centering  %图片全局居中
    \includegraphics[scale=0.5]{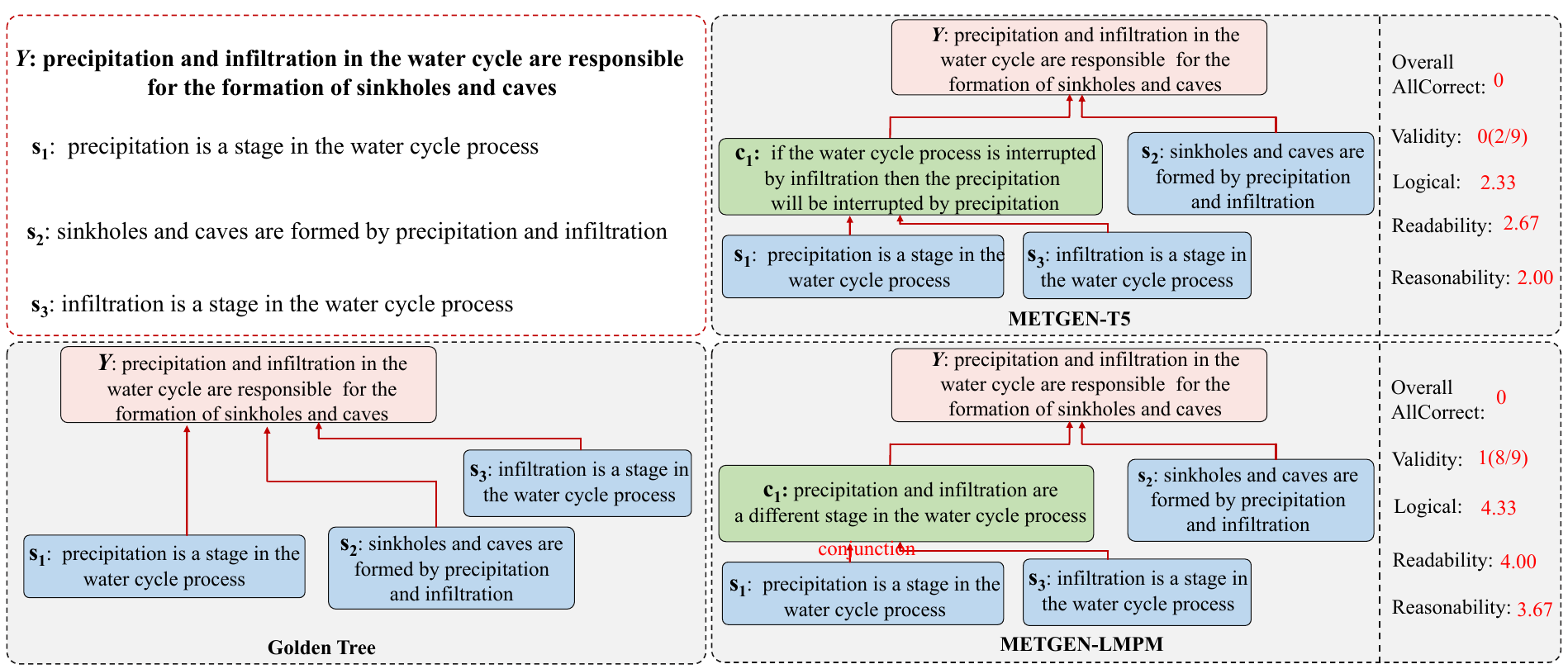} 
  \caption{Case 3: A case shows how METGEN deals with the n-premise step (n > 2), and also shows that METGEN-LMPM can effectively fuse the logical patterns and textual information from the premises for generating the conclusions in the entailment steps.}
  %\caption{The predicted entailment tree consists of two 2-premise steps, while the gold tree consists of one 3-premise step, which is allowed according to METGEN's settings. However, T5 and our proposed LMPM generate different intermediate conclusions and have different scores in the human evaluation. }
  \label{FIG:7}
\end{figure*}

In this section, we utilize three entailment trees generated by METGEN-T5 and METGEN-LMPM, which were also employed in the human evaluation for the case study. Figure~\ref{FIG:5}, \ref{FIG:6}, and \ref{FIG:7} present the task input, golden tree, trees generated by METGEN-T5 and METGEN-LMPM, along with their respective Overall AllCorrect and human evaluation results. It should be noted that Overall AllCorrect is the most stringent automatic evaluation metric. Additionally, under the Validity category, we indicate the number of evaluators (out of 9) who deemed the generated tree as valid. Furthermore, the sub-graph of METGEN-LMPM showcases the identified pattern relations.

% In this part, we use three entailment trees generated by METGEN-T5 and METGEN-LMPM, that also the three trees used in human evaluation for case study. For each tree, the task input, the golden tree, the trees generated by METGEN-T5 and METGEN-LMPM, and their corresponding Overall AllCorrect and human evaluation results\footnote{Note that Overall AllCorrect is the strictest automatic evaluation metric. And in Validity, we also show how many of the 9 evaluators consider the given generated tree is valid.} are presented, as shown in Figure~\ref{FIG:5}, ~\ref{FIG:6}, and  ~\ref{FIG:7}. We also show the types of the pattern relations identified LMPM in the sub-graph of METGEN-LMPM.

 Firstly, Figure~\ref{FIG:5} presents an illustrative example wherein LMPM outperforms T5 by generating a more accurate intermediate conclusion ($c_1$) that aligns closely with the ground truth and achieves a higher BLEURT score. Specifically, the BLEURT score for $c_1$ generated by T5 is 0.21, whereas LMPM achieves a score of 0.89. Furthermore, the metrics displayed in Figure~\ref{FIG:5} demonstrate that METGEN-LMPM outperforms METGEN-T5 in all aspects, owing to the generation of superior conclusions characterized by logical consistency during the entailment step.

% Firstly, Figure~\ref{FIG:5} gives an example showing LMPM generates a better intermediate conclusion ($c_1$) than T5, which is more in line with the ground truth and achieves a higher BLEURT score. The BLEURT score of $c_1$ generated by T5 is 0.21, while by LMPM is 0.89. In addition, the scores of METGEN-LMPM under all metrics in Figure~\ref{FIG:5} are much better than METGEN-T5 because a better conclusion with logical consistency is generated in the entailment step.
% Firstly, Figure~\ref{FIG:5} gives an example showing LMPM generates a better intermediate conclusion ($c_1$) than T5, that is more in line with the ground truth and achieves a higher BLEURT score. The BLEURT scores of $c_1$ generated by T5 is 0.21, while by LMPM is 0.89. In addition, the scores of METGEN-LMPM under all metrics in Figure~\ref{FIG:5} are much better than METGEN-T5, because the better conclusion with logically consistency is generated in the entailment step.

% Secondly, Figure~\ref{FIG:6} gives an interesting example. We observe that in METGEN-LMPM, all results of human evaluation metrics are great, but the Overall AllCorrect score of automatic evaluation is still 0. Therefore, the automatic evaluation might misjudge some valid trees and thus underestimate the performance, which is also mentioned by \citet{Dalvi2021}. In addition, compared to METGEN-T5, our proposed METGEN-LMPM accurately captures the logical relationship between premises, thereby generating logically consistent conclusion $C_1$. As a result, we have obtained improved human evaluation metrics. 

Figure~\ref{FIG:6} presents an intriguing example. Notably, in METGEN-LMPM, all the human evaluation metrics yield favorable results, while the Overall AllCorrect score of the automatic evaluation remains 0. This suggests that the automatic evaluation might erroneously disregard some valid trees, thereby underestimating the performance, as also noted by Dalvi et al \citet{Dalvi2021} Furthermore, when compared to METGEN-T5, our proposed approach, METGEN-LMPM, adeptly captures the logical relationship between premises, leading to the generation of a logically consistent conclusion, denoted as $C_1$ and $c_2$. Consequently, we observe an improvement in the human evaluation metrics. 

% Secondly, Figure~\ref{FIG:6} gives an interesting example. We observe that in METGEN-LMPM, all results of human evaluation metrics are great but the Overall AllCorrect score of automatic eveluation is still 0. Therefore, the automatic evaluation might misjudge some valid trees and thus underestimate the performance, which is also mentioned by \citet{Dalvi2021}.  
% Finally, for the golden tree consisting of the n-premise steps (n>2), METGEN allows the predicted entailment tree to have a different structure from the golden tree. As mentioned in Section 4.4, METGEN decomposes the n-premise step (n > 2) into several valid 2-premise steps before conclusion generation. In Figure~\ref{FIG:7}, the intermediate conclusion ({$c_1$}) generated by METGEN-T5 does not fuse the logical patterns and textual information from the premises ($s_1$ and $s_3$), and all evaluators consider it to be an invalid intermediate conclusion described by Validity. Furthermore, the  LMPM samples the logical pattern between the premises and uses the latent logical pattern to generate a more reasonable conclusion, which most evaluators accept. 

Lastly, concerning the golden tree comprising n-premise steps (where n > 2), METGEN allows the predicted entailment tree to exhibit a distinct structure from the golden tree. As described in Section 4.4, METGEN decomposes the n-premise step into several valid 2-premise steps prior to conclusion generation. In Figure~\ref{FIG:7}, the intermediate conclusion ($c_1$) generated by METGEN-T5 fails to integrate the logical patterns and textual information from premises ($s_1$ and $s_3$), rendering it an invalid intermediate conclusion according to the Validity criterion evaluated by all the participants. Conversely, LMPM successfully extracts the logical pattern between the premises and employs the latent logical pattern to generate a more plausible conclusion, which is generally accepted by the majority of evaluators.

% Finally, for the golden tree consisting of the n-premise steps (n>2), METGEN allows the predicted entailment tree to have a different structure with the golden tree. As we metioned in Section 4.4, METGEN decomposes the n-premise step (n > 2) into several valid 2-premise steps before conclusion generation. In Figure~\ref{FIG:7}, the intermediate conclusion ({$c_1$}) generated by METGEN-T5 does not fuse the logical patterns and textual information from the premises ($s_1$ and $s_3$), and all evaluators consider it to be an invalid intermediate conclusion described by Validity. And LMPM, which samples the logical pattern between the premises and uses the latent logical pattern to generate a more reasonable conclusion, is accepted by most evaluators. 

\end{document}